\documentclass[conference]{IEEEtran}
\usepackage{times}
\usepackage{amsmath}
\usepackage{graphicx}
\usepackage{helvet}
\usepackage{sansmath}

\usepackage[numbers]{natbib}
\usepackage{multicol}
\usepackage[bookmarks=true]{hyperref}

\pdftrailerid{}
\hypersetup{pdfauthor={}, pdftitle={}, pdfsubject={}, pdfkeywords={}, pdfcreator={}, pdfproducer={}}

\begin{document}

\title{Heed: Obedience Reward and Written\\Memory for Vision-Language-Action Policies
\unskip}

\author{\authorblockN{Mehmet Turan Yard{\i}mc{\i}}
\authorblockA{Karlsruhe Institute of Technology}}

\maketitle

\begin{abstract}
Some tasks cannot be solved from the current camera image. Take a robot that boils an egg. It has to know how long the egg has been in the water, and the image does not show this. A vision-language-action (VLA) policy has a vision-language model (VLM) and an action head, the part that outputs the actions. The VLM can keep such history as text, and several systems give the policy a written memory. A written memory reaches the action head as text, like an instruction. It helps only if the action head follows that text. We measure this on the VLA policy GR00T N1.7. An instruction given at the start of an episode selects the task. An instruction that replaces it during the episode is mostly ignored. In our test, the new instruction named another task that is feasible in the same scene. The policy completed the new task in 96 of 2160 episodes, and in 44 of these it finished the old task first. A second VLA policy, $\pi_{0.5}$, completed the new task in 18 of 2160 episodes. Our method, Heed, has two parts. The first is an obedience reward for online reinforcement learning (RL). The reward pays for fulfilling the instruction currently in the policy's input, and the instruction is switched within one scene. Reading the text is then the only way to know which goal pays. The second part is a memory of notes written by the VLM. GR00T N1.7 is released with a shortened VLM that keeps 16 of the 28 layers of its language model. Writing notes needs a complete VLM. We use a public model whose language model has all 28 layers. With it in place, we run four checkpoints of the policy, each on the group of tasks it was fine-tuned on. Together they solve 332 of 400 episodes, against 335 with the shortened VLM. We asked the VLM for notes on one late frame from each of 60 episodes, with five wordings of the prompt. With the best wording, the notes give the state of the task correctly in 34 of the 60 frames. Notes make the text input long. A 500-token note that carries no instruction lowers the success of one released checkpoint, with its shortened VLM, from 82 to 33 of 100. We fine-tuned the action head of this checkpoint with the complete VLM frozen. Fine-tuned without notes, the head drops from 86 to 54 under the same note. Fine-tuned with notes of varying length in its input, it solves 88 of 100 under the 500-token note. Our hypotheses: the obedience reward raises the share of switches the policy follows, and notes raise the success of an obedient head on memory tasks. The main evaluation of the full paper uses two household benchmark suites, LocoManiLab and RobotAgentBench; LIBERO serves only for fast preliminary tests.

\end{abstract}

\IEEEpeerreviewmaketitle

\section{Introduction}\label{sec:1}

\begin{figure*}[t]
\centering
\includegraphics{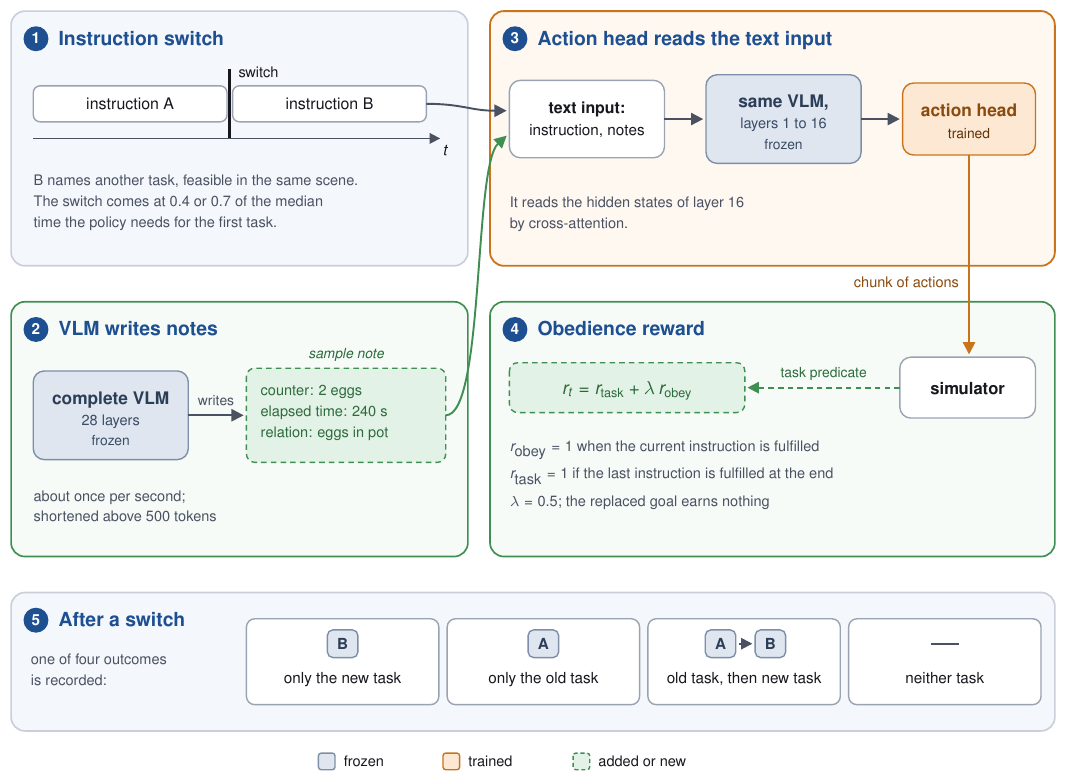}\par\vspace{\baselineskip}
\caption{Overview of Heed. (1) The instruction is switched during an episode. (2) The VLM writes notes. (3) Instruction and notes reach the action head through the same text input. (4) The reward pays for the instruction currently in the input. (5) After a switch, one of four outcomes is recorded.}
\label{fig:method}
\end{figure*}

The simplest task that needs more than the current camera image is keeping time. A soft, a medium and a hard boiled egg look exactly the same in the pot. A policy that maps the current image to an action fails at this task, at any level of training. The same holds for counting scoops, for waiting, and for putting an object back where it was.

One answer is a memory module added to the policy and trained with it. One such module keeps a store of selected frames~\cite{2510.20328}. Others keep a learned state vector, a set of numbers that people cannot interpret.

A language model keeps such information as text. It writes a count, an elapsed time and a description of the scene, and reads them back later. A vision-language-action (VLA) policy contains such a model, its vision-language model (VLM), and an action head, the part that outputs the actions. Several systems let the VLM write a memory in language~\cite{2603.03596,2602.21013,2611.30145}. Unlike a learned state vector, a written memory can be read by people. A user can see what the robot has recorded about its task and can correct a wrong entry.

Text in the input helps only a policy that acts on text. VLA policies have been found to follow the scene more than the instruction~\cite{2602.17659}. When the instruction changes during execution, they keep doing the old task~\cite{2603.17300,2611.28803}. One analysis explains why the failure survives imitation on demonstrations that carry a single instruction. After the policy has committed to a task, reading the instruction no longer changes the imitation loss~\cite{2611.34280}. Our measurement on GR00T N1.7 agrees (Section~\ref{sec:6}). We therefore treat memory and instruction following as one problem and begin with instruction following. The action head has to learn to do what the text in its input says. We teach this with reward, in online reinforcement learning (RL), from the policy's own experience.

Contributions:

\begin{enumerate}
\item An obedience reward for online RL of a VLA action head: the reward pays for the instruction currently given, and instructions are switched within one scene during the episode.
\item A memory of notes written by the VLM, delivered to the action head as text and scored against the true state.
\item An evaluation on two household benchmark suites, LocoManiLab and RobotAgentBench, that records four outcomes after a switch and separates following an instruction from using a memory.
\item Measurements on GR00T N1.7 before any training with the reward: its rate of following a switched instruction, and what note writing demands of its VLM.
\end{enumerate}

\section{Related work}\label{sec:2}

Instruction switches. The LIBERO benchmark~\cite{2306.03310} has simulated manipulation tasks for one robot arm. Its Goal suite has ten tasks in one scene. ReSteer~\cite{2603.17300} defines steerability as success after a switch of the instruction and reports 0.40 for a recent VLA policy on the Goal suite. It raises steerability by generating switch trajectories and fine-tuning by imitation. DeRail~\cite{2611.28803} builds switch benchmarks. On its LIBERO-Goal-Switch benchmark, the success of $\pi_{0.5}$ falls from 90.0\% to 34.4\% as the switch comes later. Its main method steers actions at inference without training. Wormhole~\cite{2611.34280} measures the same collapse. It produces switch data with a scripted teacher that carries the arm toward the new goal and hands control back to the policy. For $\pi_{0.5}$, imitation on this data raises success from about 1\% to between 85\% and 92\%. The gain is for switches made after the policy has committed to its first task. LIBERO-CF~\cite{2602.17659} assigns alternative feasible instructions in the same scene and corrects the policy at inference. These works change the data or the inference procedure. Changing the data needs a procedure that generates switch trajectories. We change the reward, which needs only the outcome.

Rewards tied to the instruction. Learning from Hindsight~\cite{2607.09042} runs online RL. A rollout is an episode the policy has just played. A separate large VLM relabels a failed rollout from each low-reward group with the instruction that it fulfilled. The instruction thus changes after the episode. Outside VLA policies, MAVIC~\cite{2605.12655} studies multi-agent RL in which instructions switch during an episode, and it rewards the current instruction. It shows that, at a switch, the estimate of future reward mixes the returns of different instructions. Correcting the target of that estimate prevents this. We bring the switched-instruction reward to the action head of a VLA policy.

Written memory. MEM~\cite{2603.03596} trains a language memory with supervised targets from a language model. Notes-to-Self~\cite{2602.21013} trains a scratchpad by supervision. RMR-VLA~\cite{2611.30145} uses online RL with a reward for how well the subgoal field of a written memory matches labels. Gamma~\cite{2610.12793} checks each line of a VLM's written memory against an object detector's output on the frames the VLM saw. It finds that its writer reports progress without evidence in the frames. MemER~\cite{2510.20328} reports that a text memory performs below a memory of selected frames. In 2AM~\cite{2609.11308}, a separate agent writes guidance, and an action model is trained by imitation to act on it. In HeLM~\cite{2611.02660}, a separate VLM keeps the task history as text and updates it when it detects an event from a fixed list. In MemER, Gamma and HeLM, the acting policy receives a subtask or an instruction that another model derives from the memory. In MEM, one model fills both roles. These works assume that the acting policy follows the text it receives. We measure whether it does, and we train it by reward.

\section{Problem statement}\label{sec:3}

The VLM of the policy reads images and text. From the VLM's output, the action head produces a chunk of actions. A chunk is a short sequence of actions executed before the next pass of the policy. We call one pass of the policy a call. The task reward is sparse: 1 at success, 0 otherwise.

Obedience. We call an action head obedient when a change of the text in its input changes its behaviour to match the new text. We measure it with an instruction switch. During an episode, the instruction is replaced with one for another task that is feasible in the same scene. We record one of four outcomes: the policy completes only the new task, only the old task, the old task and then the new one, or neither.

Memory tasks. In a memory task, the observation at one step does not determine the right action. The policy needs information from earlier steps: a count, an elapsed time, an earlier position.

We ask two questions. How does an action head become obedient through its own experience? Can an obedient head then solve memory tasks from notes in its text input, without a separate memory mechanism?

\section{Method}\label{sec:4}

\subsection{Base}\label{sec:4.1}

We use GR00T N1.7, a later version of GR00T N1~\cite{2503.14734}. The instruction reaches the action head only through the hidden states of the VLM. These are the vectors that the VLM's layers output. The action head reads the hidden states of layer 16 of the VLM's language model. It reads them by cross-attention. The action head produces actions by sampling, starting from random noise. It processes a sequence that holds the robot state and the noisy actions. Each element of that sequence takes in a weighted mix of those hidden states. Training follows the public RL recipe for this model~\cite{rlinf}. The algorithm is proximal policy optimisation~\cite{1707.06347}. It raises the probability of actions whose return was higher than expected. For exploration, the recipe adds further noise during the sampling. The recipe updates the action head and a value head, a small network that predicts the expected return. In our runs the VLM receives no gradient.

\subsection{Obedience reward}\label{sec:4.2}

\begin{equation}
r_t = r_{\mathrm{task}} + \lambda\, r_{\mathrm{obey}}, \quad \text{with } \lambda = 0.5.
\end{equation}

Here $r_{\mathrm{task}}$ is the task reward, $r_{\mathrm{obey}}$ is the obedience term and $\lambda$ is the weight of the obedience term. Both terms are sparse. The term $r_{\mathrm{obey}}$ is 1 at the step where the instruction currently in the input is fulfilled, and 0 otherwise. It pays once per instruction. The term $r_{\mathrm{task}}$ is 1 if the last instruction is fulfilled at the end of the episode. In an episode with one fixed instruction, both terms pay for the same goal. After a switch to another task, both terms pay for the new goal. Reaching the replaced goal earns nothing. A task can also be given as several instructions in turn. Then $r_{\mathrm{obey}}$ pays each of them and $r_{\mathrm{task}}$ pays the last.

In simulation the reward comes from the task predicate, the simulator's own test of whether a goal holds. On a real robot, a judge that decides fulfilment from images would take its place. Section~\ref{sec:6} reports a first test of the VLM as this judge.

Contrastive instructions. Episodes are sampled in scenes where several goals are feasible. The instruction selects one goal, and only that goal pays. Ordinary RL in such a scene already has this property. The new element is the switch during the episode. A head that ignores the text after a switch collects reward by chance only.

Switch time. We switch at 0.4 and at 0.7 of the first task's median completion time: one point in the first half of the task and one in the second, set relative to completion time so that tasks of different length are interrupted at a comparable stage. At both points the first task is still unfinished in every control episode (a goal holds in 0 of 1080 at each point, for both models), so following the new instruction is still possible. The later the switch, the more often the policy completes only the old task: the share rises from 0.52 at realised fractions 0.3 to 0.5 to 0.81 at 0.7 to 0.9 for GR00T N1.7, and from 0.28 to 0.39 for $\pi_{0.5}$, while the share that completes only the new task stays below 0.04 in these bins. The early switch falls while the policy is still approaching the first goal. The late switch falls when it is close to finishing.

Instruction variety. The switches cover four families. A change of goal names another task. A negation has the form ``do not ...''. A sequence of two steps has the form ``first ..., then ...''. A correction of an action already taken has the form ``stop, put it back''. Four families are used because a head trained on one family can learn to read only the name of the goal object.

\subsection{The VLM as writer}\label{sec:4.3}

The VLM released with GR00T N1.7 keeps 16 of the 28 layers of its language model, which is enough for the action head. Writing text needs all layers. Asked a yes-or-no question, the shortened VLM puts 0.08 of its answer probability on the two words. We therefore replace it with Cosmos-Reason2-2B~\cite{cosmosreason2}, a public model built on Qwen3-VL-2B-Instruct. In the weights we loaded, its language model has 28 layers, and the first 16 have the same shapes as those of the shortened VLM. We call it the complete VLM. The action head keeps reading layer 16, so the path from observation to action has the same shape as before. The remaining layers serve text generation.

\subsection{Notes as memory}\label{sec:4.4}

About once per second the VLM writes notes. They hold counters, the elapsed time, relations between objects and the current sub-instruction. The sub-instruction is the next step toward the task. The time comes from a clock. The clock starts when the robot is switched on, and the model receives its reading as a number. When the notes exceed a budget of 500 tokens, the VLM rewrites them in shorter form. Spatial content can also be kept as a small top-down sketch that the VLM receives back as an image. The experiments in this paper use text only. The action head receives the notes through the instruction input. Every note is logged, and its counter, elapsed time and relations are scored against the true state of the simulator.

The action head has to tolerate a long text input before it can use one (Section~\ref{sec:6}). Before RL, it is therefore fine-tuned on demonstrations with notes in its input, in two stages. The first stage uses neutral notes. A neutral note repeats one sentence up to the length of the budget. The sentence says that the note carries no instruction. The second stage uses oracle notes. They have the same fields as the VLM's notes and are written from the simulator state.

\section{Hypotheses and evaluation}\label{sec:5}

Setting. GR00T N1.7 with the complete VLM, in simulation. The main evaluation uses two benchmark suites. LocoManiLab~\cite{2610.30876} has 234 household tasks for the Unitree G1 humanoid, which walks and manipulates; its kitchen scene alone holds 58 of them. RobotAgentBench~\cite{2610.34649} has household tasks in which a mobile robot with two arms must also find its tasks and recover from failures, and its scene is not reset between tasks. For the full paper we run both in Isaac Sim and port LocoManiLab to it. Obedience is trained and measured with instruction switches between tasks that are feasible in the same scene of these suites. In the released task files of LocoManiLab we count 1,648 ordered pairs of distinct atomic tasks that share a scene instance. Switches of the form ``first ..., then ...'' use their tasks with several parts. Negations and corrections are built on the same scenes. Memory is measured on tasks that we build in these scenes and that need information from earlier steps: counting, waiting and putting an object back where it was. LIBERO (Section~\ref{sec:6}) serves only for fast preliminary tests, and the full paper adds no further LIBERO experiments.

Conditions. (a) RL with the task reward and a fixed instruction. (b) RL with the obedience reward and switched instructions. (c) Condition (a) with oracle notes. (d) Condition (b) with oracle notes. (e) Condition (b) with notes from the VLM. This is the full method, Heed. (f) Condition (e), with only the sub-instruction line of each note given to the action head. Each condition is trained with three seeds. An ablation varies the weight $\lambda$.

H1 (main). After a switch, the policy completes the new instruction in a higher share of episodes in condition (b) than in condition (a).

H2 (retention). In episodes with a fixed instruction, task success in condition (b) is at least as high as in condition (a).

H3. Given oracle notes, an obedient head solves memory tasks at a higher rate than the same head without notes: (d) against (b). A head trained without switches gains less from the same notes: (c) against (a).

H4. With notes written by the VLM, condition (e), success on memory tasks approaches that of condition (d). The scored accuracy of the notes accounts for the remaining gap. Condition (f) tests whether the head needs the whole note or only the VLM's sub-instruction.

If a hypothesis fails. If H1 fails, this action head needs more than reward to read its text input. The path from text to action then has to change, and the VLM is the first candidate. If H2 fails, the policy gains obedience at the cost of task success. Either $\lambda$ is too large, or the value estimate mixes returns at the switch, as Section~\ref{sec:2} describes~\cite{2605.12655}. If H1 holds and H3 fails, using a memory needs more than following an instruction. If H3 holds and H4 fails, the content of the VLM's notes is the limit. The scored notes then show which kind of content is wrong.

\section{Preliminary measurements}\label{sec:6}

These are fast preliminary tests, on LIBERO unless stated; the main evaluation is on the two suites of Section~\ref{sec:5}.

Instruction switch. We ran the GR00T N1.7 checkpoint fine-tuned on LIBERO-Goal on 12 ordered pairs of tasks, a first task and a new task. We repeated the evaluation three times on the same pairs, with 720 episodes each time. Half of the episodes switched at 0.4 of the first task's median time and half at 0.7. Table 1 gives the four outcomes of Section III for each switch time. A policy that follows the switch would fill the first row; the second row, only the old task, is the largest. The policy completed the new task in 96 episodes (0.044), and in 44 of them it had finished the old task first. A later switch is followed less often, and an early switch often leaves neither task done. A sweep over six switch times uses the same checkpoint and pairs, with 360 episodes per switch time on one set of placements. The rows at 0.4 and 0.7 are one of the three repetitions above. At 0.1 of the median time the policy completes only the new task in 123 of 360 episodes and only the old task in 57. At 0.25, 0.4, 0.55, 0.7 and 0.85 it completes only the new task in 15, 10, 15, 4 and 0 episodes, and only the old task in 143, 181, 199, 261 and 322. Only the switch at 0.1 is followed in about a third of the episodes; from 0.25 on, this share is below 0.05. With the instruction given from the first step, this checkpoint solves 81 and 82 of 100 Goal episodes in two runs (below). In an independent evaluation, two GR00T N1.7 checkpoints reach 92\% and 96\%~\cite{2610.41609}. The failure is specific to a switch during execution.

A second model. We ran the same test with $\pi_{0.5}$~\cite{2504.16054}, a second VLA policy, using a public checkpoint fine-tuned on LIBERO~\cite{rlinf}. The pairs, the switch times and the number of episodes are the same. With a fixed instruction, $\pi_{0.5}$ solves 561 of 630 episodes of the tasks in these pairs. After a switch it completes the new task in 18 of 2160 episodes. It completes only the old task in 288 and 343 of 1080 episodes at the two switch times, and neither task in 783 and 728. $\pi_{0.5}$ also fails to follow the switch, and it more often ends with neither task done. In the original MuJoCo release of LocoManiLab, with the GR00T N1.7 checkpoint that its authors fine-tuned on their atomic training tasks, we replaced ``close the fridge door in front of you'' by ``open the cabinet door in front of you'' at one third of the episode limit. In 10 episodes of this one pair, the policy completed only the old task in 8 and neither task in 2.

\begin{table}[t]
\renewcommand{\thetable}{\arabic{table}}\setcounter{table}{0}
\centering
\caption{Outcomes after an instruction switch, 1080 episodes per switch time.}
\label{tab:switch}
\begin{tabular}{lcc}
\hline
 & Switch at 0.4 & Switch at 0.7\\
\hline
Only the new task & 40 & 12\\
Only the old task & 541 & 786\\
Old task, then new task & 20 & 24\\
Neither task & 479 & 258\\
\hline
\end{tabular}
\end{table}

Replacing the VLM. All 176 tensors in the 16 language-model layers of the shortened VLM differ from their counterparts in the complete VLM. The median relative difference, the norm of the difference over the complete tensor's norm, is 0.024. Replacing the VLM therefore changes the input of the action head. The authors of GR00T N1.7 released four checkpoints, each fine-tuned on one LIBERO suite. In each checkpoint, we replaced every VLM tensor with its counterpart from the complete VLM and left the action head unchanged. We ran each on its own suite. The suites vary the layout (Spatial), the objects (Object), the goal within one scene (Goal) and the length of the task (Long). The policy runs with each VLM on the same placements of the objects. Out of 100 episodes, the complete VLM gives 90 on Spatial, 88 on Object, 83 on Goal and 71 on Long. The shortened VLM gives 95 on Spatial, 85 on Object, 81 on Goal (82 in a second run) and 74 on Long. Pooled, this is 332 against 335 of 400. The two VLMs disagree on 77 placements: 40 are solved only with the shortened VLM and 37 only with the complete one. The exact McNemar test for paired outcomes gives p = 0.82.

Fine-tuning with the complete VLM. We fine-tuned the action head of the Goal checkpoint for 200 steps on the Goal demonstrations, with the complete VLM frozen. It solves 86 of 100 episodes, against 83 before this fine-tuning on the same placements. For the 86, the 95\% Wilson score interval, a confidence interval for a success rate, is 0.78 to 0.91. The two disagree on 13 placements, 8 solved only after this fine-tuning and 5 only before it (exact McNemar test, p = 0.58).

Notes. We took one late frame from each of 60 episodes of the Spatial, Object and Goal suites, 30 successful and 30 failed. For a failed episode it is the last saved frame, at 7.2 s or 11.2 s. On each frame the complete VLM wrote notes with the clock given. We asked for the notes with five wordings of the prompt. One wording copies the elapsed time correctly in all 60 notes. The state of the task, written in the relation lines of the note, is harder. The best wording gives it correctly in 34 of 60 frames (95\% Wilson score interval 0.44 to 0.68). Two of the other wordings give it correctly in 15 of 60. With the best wording, 59 notes give a counter. In 6 of them the counter contradicts the relations written in the same note. On 60 late and 60 first frames of recorded LocoManiLab episodes the same test is inconclusive: two wordings state that the task is complete on 118 and 110 of the 120 frames.

The VLM as judge. On 200 episodes of the Long checkpoint, half of them successful, each VLM saw two camera images of the last step. It was asked whether the robot had completed the task and told to answer yes or no. We formed pairs of one successful and one failed episode. For the complete VLM, the successful episode gets the higher probability of ``yes'' in 0.52 of the pairs (95\% interval 0.44 to 0.60, bootstrap over episodes). Chance is 0.5. For the shortened VLM the share is 0.62 (0.54 to 0.70), although it puts only 0.08 of its probability on yes and no (Section~\ref{sec:4.3}). Both are too weak to replace the predicate.

A long text input. The original LIBERO-Goal checkpoint, with its shortened VLM, solves 33 of 100 episodes when a neutral note of 500 tokens is added to the instruction. The processor sets no length limit on the text, so the note is not cut. Without the note, it solves 82 of 100 on the same placements. The Goal checkpoint fine-tuned with the complete VLM solves 54 of 100 under the same note of 500 tokens. This checkpoint differs from the original one in two ways, its VLM and its fine-tuning. Either can cause the gap between 54 and 33.

Fine-tuning with notes in the input. We repeated the 200-step fine-tuning of the Goal checkpoint with a neutral note added to the instruction of each demonstration. The length of each note was drawn at random from 0, 250 and 500 tokens. Table 2 compares the resulting checkpoint with the checkpoint fine-tuned without notes, on the same placements. Without notes in training, success falls as the note grows, from 86 to 54. With notes in training, it stays between 83 and 88. On one NVIDIA V100 GPU, with a batch of 640, the median training step takes 212 s with notes and 78 s without. Both fine-tunings ran with one training seed. Whether the action head uses the content of a note is the question of H3.

\begin{table}[t]
\renewcommand{\thetable}{\arabic{table}}\setcounter{table}{1}
\centering
\footnotesize
\setlength{\tabcolsep}{4pt}
\caption{Success out of 100 episodes with a neutral note added to the instruction. Columns: note of 0, 250 and 500 tokens, where 0 means no note.}
\label{tab:2}
\begin{tabular}{lccc}
\hline
 & 0 & 250 & 500\\
\hline
\parbox[t]{0.30\linewidth}{\raggedright\strut Fine-tuned with notes\strut} & 85 & 83 & 88\\
\parbox[t]{0.30\linewidth}{\raggedright\strut Fine-tuned without notes\strut} & 86 & 73 & 54\\
\parbox[t]{0.30\linewidth}{\raggedright\strut Exact McNemar test between the two\strut} & p = 1.00 & 0.076 & 1.4e-7\\
\hline
\end{tabular}
\end{table}

The time of a long input. On one NVIDIA V100 GPU, one call of the policy takes 357, 422, 510, 680 and 1002 ms for a note of 0, 100, 250, 500 and 1000 tokens.

\section{Limits}\label{sec:7}

Every test on tasks in this paper is in simulation, and these tests are preliminary.

\section*{Use of language models}
Large language models were used to draft and edit the text of this paper from the author's written notes and answers, and to write and review code. The author read and checked all content and is responsible for it.

\bibliographystyle{plainnat}
\bibliography{references}

\end{document}